\documentclass[letterpaper, 10 pt, conference]{ieeeconf}  % Comment this line out if you need a4paper

\IEEEoverridecommandlockouts                              % This command is only needed if 
\usepackage{graphicx} % for pdf, bitmapped graphics files
\usepackage{xcolor}
\usepackage[noadjust]{cite} % This gives citation ranges
\let\proof\relax 
 \let\endproof\relax
\usepackage{amsmath,amsthm} % assumes amsmath package installed
\usepackage{amssymb}  % assumes amsmath package installed
\usepackage{subfiles}
\usepackage[caption=false]{subfig}

\newtheorem{thm}{Theorem}
\newtheorem{asu}{Assumption}
\newtheorem{obs}{Observation}
\newenvironment{proofs}{%
	\proof}{\endproof}

\usepackage{mathtools}
\mathtoolsset{showonlyrefs} % only equations that are referenced will be numbered

\title{\LARGE \bf
Non-Uniform Robot Densities in Vibration Driven Swarms Using Phase Separation Theory$^\dag$*%\titlemarker
}

\author{Siddharth Mayya$^{1}$, Gennaro Notomista$^{1}$, Dylan Shell$^{2}$, Seth Hutchinson$^{1}$, and Magnus Egerstedt$^{1}$% <-this % stops a space
\thanks{$^\dag$ This work has been submitted to the IEEE for possible publication. Copyright may be transferred without notice, after which this version may no longer be accessible.}
\thanks{*This work was supported by the U.S. Office for Naval Research under Grant N0014-15-1-2115.}% <-this % stops a space
\thanks{$^{1}$S. Mayya, G. Notomista, S. Hutchinson, and M. Egerstedt are with the Institute for Robotics and Intelligent Machines, Georgia Institute of Technology, Atlanta, GA, USA {\tt\small \{siddharth.mayya,g.notomista, seth,magnus\}@gatech.edu}}%
\thanks{$^{2}$D. Shell is with the Department of Computer Science and Engineering at Texas A\&M University, College Station, TX, USA
	{\tt\small dshell@tamu.edu}}%
}

\begin{document}

\maketitle
\thispagestyle{empty}
\pagestyle{empty}

%%%%%%%%%%%%%%%%%%%%%%%%%%%%%%%%%%%%%%%%%%%%%%%%%%%%%%%%%%%%%%%%%%%%%%%%%%%%%%%%

\begin{abstract}
In robot swarms operating under highly restrictive sensing and communication constraints, individuals may need to use direct physical proximity to facilitate information exchange. However, in certain task-related scenarios, this requirement might conflict with the need for robots to spread out in the environment, e.g., for distributed sensing or surveillance applications. This paper demonstrates how a swarm of minimally-equipped robots can form high-density robot aggregates which coexist with lower robot densities in the domain. We envision a scenario where a swarm of vibration-driven robots---which sit atop bristles and achieve directed motion by vibrating them---move somewhat randomly in an environment while colliding with each other. Theoretical techniques from the study of far-from-equilibrium collectives and statistical mechanics clarify the mechanisms underlying the formation of these high and low density regions. Specifically, we capitalize on a transformation that connects the collective properties of a system of self-propelled particles with that of a well-studied molecular fluid system, thereby inheriting the rich theory of equilibrium thermodynamics. This connection is a formal one and is a relatively recent result in studies of motility induced phase separation; it is previously unexplored in the context of robotics. Real robot experiments as well as simulations illustrate how inter-robot collisions can precipitate the formation of non-uniform robot densities in a closed and bounded region.
%For a team of differential drive bristle-actuated robots, we show that
%spontaneous phase separation can occur, which can be usefully applied in many
%areas. In general, a vast literature in active particles can be tapped into.
\end{abstract}
%%%%%%%%%%%%%%%%%%%%%%%%%%%%%%%%%%%%%%%%%%%%%%%%%%%%%%%%%%%%%%%%%%%%%%%%%%%%%%%%
\section{Introduction} \label{sec:intro}

% Propose: flip the story around, switch the ordering, to put emphasis on
% robotics first (phase-separation as something with task utility up-front),
% then build up to Stat. Mech.; emphasize macroscopic language, via classical
% theory connects for active particles.
% 
% Application example, sampling, requires different densities.
Swarm robotic systems are comprised of robots that, though individually simple, aim to produce useful group-level behaviors via local interactions (see surveys \cite{csahin2004swarm,brambilla2013swarm} and references within). For such swarm systems, a significant research focus has been on achieving emergent and self-organized collective behaviors via strictly local rules, e.g. \cite{bonabeau99swarm}. \par
Designing collective behaviors becomes especially challenging when considering scenarios where robots with limited sensing and communication capabilities perform tasks which require them to spread across an environment, e.g., for distributed sensing or environmental surveillance applications \cite{low11active,elston09toward}. Under these circumstances, the robots might require spatial proximity to facilitate information exchange among themselves, while simultaneously performing task related actions \cite{tovar2007using,diaz2013multi}.\par
In this paper, we highlight a mechanism to achieve coexisting regions of low and high robot density in swarms with severe constraints on sensing and communication. We demonstrate these behaviors on a team of vibration-driven robots, called \emph{brushbots} \cite{notomista2019brushbot}, which achieve directed locomotion by vibrating bundles of flexible bristles. In particluar, these robots do not possess sensors to detect other robots and simply traverse the environment while colliding with other robots. We illustrate that the mechanisms underlying the obtained density distributions can be explained using results from statistical mechanics, which investigates how macroscopic phenomena observed in physical systems can be related to the microscopic behaviors of constituent particles~\cite{chandler1987introduction}. \par
While equilibrium statistical mechanics provides an extensive vocabulary to describe macroscopic behaviors, much of the classical theory deals with idealized interactions among particles, limiting its applicability in swarm robotic systems \cite{spears05physicomimetics}. However, the study of physical systems that are far from equilibrium has generated a great deal of recent excitement~\cite{nas2010}, given its ability to systematically analyze complex collectives in nature \cite{bechinger2016active}. In these \emph{active matter} systems, an interplay between self-propulsion, inter-particle effects, and environmental forces leads to a wide-variety of emergent behaviors~\cite{vicsek1995novel,marchetti2013hydrodynamics}. This paper takes advantage of a lesser-known \emph{formal connection between certain types of active matter systems and equilibrium thermodynamics} to develop a microscopic description for a team of self-propelled brushbots, while retaining the extensive benefits of the classical thermodynamic theory.\par  %systems as diverse as fish schools~\cite{katz2011inferring},
%colloidal-suspensions~\cite{kummel2013circular}, and bacterial colonies~\cite{berg2008coli}. 

We envision a team of brushbots moving somewhat randomly in a closed domain, while colliding with each other. The simultaneous formation of regions with lower and higher robot density is intuitively supported by two observations. Firstly, a given robot's speed decreases with increasing robot density around it---a direct consequence of the inter-robot collisions experienced by the robot. Secondly, a system of particles---which are embodied by robots in this context---tend to accumulate in regions where they move more slowly \cite{schnitzer1993theory}. Such a phenomenon is known as \emph{motility induced phase separation} (MIPS) in the physics literature \cite{cates2015motility} and allows the formation of single or multiple high-density robot clusters which contain only a subset of the total robots in the domain. Recently established connections between such observations in active particle systems and equilibrium thermodynamics \cite{cates2015motility} allow us to impose design constraints on the motion characteristics of the robots displaying such behaviors. \par 

A team of brushbots provides an ideal platform for achieving such variable density behaviors, since their minimalist construction makes them robust to the force experienced during inter-robot collisions even at relatively high speeds \cite{notomista2019brushbot}. Additionally, the inherently noisy dynamics of the brushbots ensures that---similar to active matter systems---high-density robot clusters do not last infinitely long. \par 

The outline of the paper is as follows.  The next section makes more detailed connections to relevant literature.  Section~\ref{sec:models} introduces the stochastic differential equation describing the dynamics of each brushbot, and uses existing results on inter-robot collisions to derive a density dependent speed profile for each robot. In Section~\ref{sec:mips}, this model is leveraged to discuss the conditions under which motility induced phase separation can occur by drawing connections with an equivalent system of particles at thermal equilibrium. Simulations confirm the formation of robot aggregations for varying parameter ranges.  In Section~\ref{sec:exp}, the mechanism is deployed on a team of real differential-drive--like brushbots to illustrate the formation of high and low robot densities.  Section~\ref{sec:conc} concludes the paper.

\section{Related Work} \label{sec:lit}
%Heiko Hamann papers, clustering in swarm robotics.

\subsection{Aggregation in Swarm Robotics}

Within the robotics literature, several threads of work have examined how to get robots, under a variety of circumstances, to aggregate in certain places or at particular densities~\cite{trianni2003evolving,soysal2005probabilistic,ren2005survey,garnier2005aggregation}. The problem of aggregation becomes especially relevant when the robots have basic sensing capabilities and limited computational resources, since physical proximity is then essential to enable more sophisticated swarm behaviors, e.g., \cite{mayya2018localization}. In \cite{gauci14self}, the authors investigate the ability of robots with access to minimal but unrestricted range information to aggregate. \cite{chen12seg} investigates how robots can achieve different packing arrangements based on the concept of different radii of interaction. \par 
In contrast to the above methods which cluster robots into high-density groups, this paper investigates the maintenance of two different densities which must coexist over time. Thus, our approach emphasizes the notion of different phases, with distinct densities, for which intermediate densities are dynamically unstable and hence vanish. We demonstrate that the physical effects of inter-robot collisions cause the robots to slow down, precipitating the formation of such regions.

%In such situations, Then the work becomes relevant for the broader class of robot swarms where physical proximity is essential to facilitate information exchange and
%enable more sophisticated swarm level behaviors (see~\cite{mayya2017collisions}). Both~\cite{chen12seg,gauci14self} are notable minimalist examples of aggregation for robots in this class, though their approach involves producing different packing arrangements based on the concept of different radii of interaction. Our approach instead emphasizes the notion of different phases, with distinct densities, for which intermediate densities are dynamically unstable and hence vanish.

%As simple robots, bristle-driven brushbots~\cite{notomista2019brushbot} are an
%ideal platform for testing aggregation and density adjustment mechanisms
%for two reasons.  Firstly, the inherently noisy dynamics of the
%brushbots ensures that\,---similar to active matter systems---\,high-density
%robot clusters never last forever.  Secondly, the simplistic
%construction of these robots makes them robust to the forces experienced during
%inter-robot collisions even at relatively high
%speeds~\cite{notomista2019brushbot}.  
%For such a swarm of bristle-bots
%operating in a confined environment, we develop an analytical model for the
%average speed of each robot as a function of the robot density.  This allows us
%to characterize the aggregation properties of the swarm in terms of the
%dynamical parameters of the individual robots. 

\subsection{Active Matter Systems}
%As mentioned in Section \ref{sec:intro}, using idealized equilibrium  as models of
%robots is problematic because they mismatch many of the aspects inherent to robots
%(mechanical considerations, finite and erratic interaction range,
%stochasticity, etc.), while further imposing somewhat singular restrictions
%(e.g., typically interaction laws will conserve energy and momentum). \par 

\emph{Active matter physics} deals with the study of self-driven particles which convert stored or ambient energy into organized movement, and can be used to describe systems as diverse as fish schools, colloidal-suspensions, and bacterial colonies (see surveys ~\cite{marchetti2013hydrodynamics,bechinger2016active} for a discussion on collective phenomena in active matter systems).

In particular, the study of \emph{motility induced phase separation}
focuses on how active matter systems consisting of self-propelled particles under purely repulsive interactions can spontaneously phase separate. 
In particular, it has been shown that, under suitable modeling assumptions, a direct connection can be made between such a system and a passive simple fluid at equilibrium with attractive interactions among the molecules. Equilibrium systems like these are well understood~\cite{atkins2011physical}, thus allowing a large body of analysis to be transferred for the study of active self-propelled systems, as discussed in~\cite{cates2015motility,stenhammar2013continuum,fily2014freezing}. In this paper, we leverage this connection to demonstrate how a swarm of brushbots \cite{notomista2019brushbot} with severely constrained sensing capabilities can phase separate into regions of high and low robot density.
\begin{comment}
Many interesting phenomena in equilibrium thermodynamics emerge from varying interaction potentials among robots. But implementing interaction potentials are difficult with robots required sensors and the ability to simulate those forces. But with our current work we can equate the robots with a virtual system in equilibrium without the need for any sensing -- just density induced collisions among the robots. 
, as well as high-density clustering~\cite{redner2013reentrant}. 
\end{comment}

\section{Motion and Collision Model} \label{sec:models}
\subsection{Motion Model} \label{sec:mm}
As introduced in Section \ref{sec:intro}, the \emph{brushbot} \cite{notomista2019brushbot} does not posses sensors to detect other robots and is robust to collisions, making it an ideal platform on which to study the formation of collision-induced variable density aggregates. In this section, we first briefly motivate the dynamical model for this robot operating in an environment without the presence of other robots. Following this, we analyze the effects of interactions between multiple robots and characterize the velocity of the robots as a function of swarm density.\par
The differential-drive--like construction of the brushbots allows the motion of the robot to be expressed using unicycle dynamics with linear and angular velocity as the inputs. For a team of $N$ robots, let $z_i = (x_i,y_i)$ and $\theta_i$ denote the position and orientation of robot $i \in \{1,\ldots,N\} \triangleq \mathcal{N}$, respectively. Each robot has a circular footprint of radius $r$ and operates in a closed and bounded domain $\mathcal{V} \subset \mathbb{R}^2$.  \par 
\begin{comment}
\begin{figure}
	\includegraphics[width=0.45\textwidth]{figs/brushbot_diff_drive.jpg}
	\caption{A differential-drive brushbot \cite{notomista2019brushbot}, actuated using two vibration motors and moving on two brushes. We model this robot using unicycle dynamics, with white noise added to the position and orientation of the robot. The brushbot provides an ideal platform to test decentralized aggregation behaviors due to its robustness to inter-robot collisions as well as stochastic dynamics, which ensures that large aggregates can naturally dissolve over time.}
	\label{fig_bbot}
\end{figure}
\end{comment}
\par
\begin{figure} 
	\includegraphics[trim={8cm 0.2cm 10cm 2cm},clip,width=0.40\textwidth]{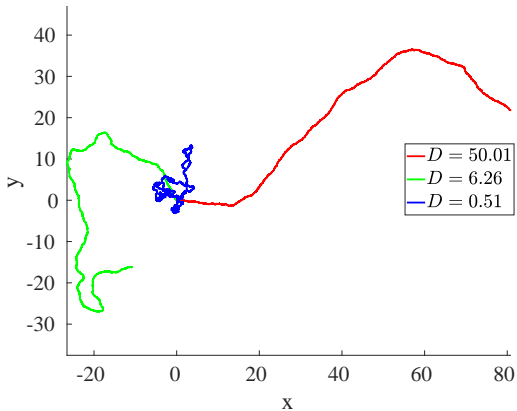}
	\caption{Illustration of the effect of diffusivity parameters on the motion of a simulated brushbot obeying the Langevin dynamics described in \eqref{eqn_dyn}. As can be seen, the effective diffusion coefficient defined in \eqref{eq:eff_diff} is indicative of the mean-square displacement of the robot over time. All three trajectories correspond to the same time horizon.}
	\label{fig:activity}
\end{figure}

 
As discussed in \cite{notomista2019brushbot}, the presence of manufacturing differences as well as the nature of bristle-based movement implies that the motion of the brushbots is somewhat stochastic. This can be modeled in the form of additive noise terms affecting the translational and rotational motion of the robots, with diffusive coefficients $D_t$ and $D_r$, respectively. Under this model, the state $(z_i,\theta_i)$ evolves according to standard Langevin dynamics, which is a stochastic differential equation, given componentwise as
\begin{align} \label{eqn_dyn}
& dx_i = v_0\cos{\theta_i} + \sqrt{2D_t}\Delta W_x, \nonumber \\
& dy_i = v_0\sin{\theta_i} + \sqrt{2D_t}\Delta W_y, \\
& d\theta_i = \sqrt{2D_r}\Delta W_\theta, \nonumber
\end{align}
where $v_0$ is the constant self-propelled speed of the robot and $\Delta W_x, \Delta W_y, \Delta W_\theta$ denote Wiener process increments representing white Gaussian noise. The total distance traveled by the robot will depend on the diffusion parameters $(D_t,D_r)$ and the speed $v_0$. The effective diffusion coefficient of the robot \cite{bechinger2016active}, measured on time scales longer than the time required for the orientation of the robot to uncorrelate (defined as $\tau_r=D_r^{-1}$), can be quantified as,
\begin{equation} \label{eq:eff_diff}
D = \frac{v_0^2}{2D_r} + D_t.
\end{equation}
Figure \ref{fig:activity} shows the trajectories of a robot for different values of the diffusivity parameter, $D$, illustrating its impact on the displacement of the robot over time. 
%All trajectories correspond to the same time horizon of simulation. % (Redundant, mentioned in the caption.)
 \par 
An important predictor of phase separation behavior in interacting systems is the \emph{activity} parameter, e.g., \cite{fily2014freezing,stenhammar2013continuum}, which we define here as
\begin{equation} \label{eqn_activity}
\mathcal{A} = \frac{v_0}{2rD_r}.
\end{equation}
The activity parameter is proportional to the distance traveled by a robot before its direction uncorrelates completely \cite{stenhammar2013continuum}, and will be used to characterize the proportion of high and low robot density regions in Section \ref{sec:mips}. In Section \ref{sec:exp}, we use robotic experiments to empirically determine the diffusion coefficients $D_t$ and $D_r$ of the brushbot, but for now, we simply assume that these parameters are available to us and are constant throughout the swarm.

\subsection{Inter-Robot Interaction Models} \label{sec:col}
In the previous section, we described the dynamics of a single robot moving with a self-propelled speed $v_0$ through the domain. Given a team of $N$ brushbots moving randomly according to these dynamics in the domain $\mathcal{V}$, we now develop a model to describe the effects of inter-robot collisions on the average speed of the robots. \par 
For a team of colliding robots, average robot speeds reduce with increasing density in the region around the robot---a direct consequence of the density-dependent collision rates~\cite{mayya2018localization}. The distribution of robots over the domain can be described in terms of the ``coarse-grained'' density $\lambda: \mathcal{V} \rightarrow \mathbb{R}_+$, obtained by overlaying a suitable smoothing function, e.g. \cite{stenhammar2013continuum}, over the microscopic density operator,
\begin{equation} \label{eqn_mic_den}
\sum_{k=1}^{N} \delta(z - z_k),
\end{equation}
defined at each point $z \in \mathcal{V}$. Here $\delta$ denotes the Dirac delta function. We defer the actual computational details of the coarse-grained density to Section \ref{sec:sim}, which discusses the simulation results. \par 
A given robot $i$ will experience varying collision rates depending on the density of robots surrounding it. The developed model is agnostic to which robot we pick since the following mean-field analysis applies to any robot in the domain \cite{mayya2017collisions}. Then, the speed of a given robot $i$ is a function of its location as well as the robot density over the domain $\lambda$. This dependence is formalized as $v(g(\lambda,z_i)) \in \mathbb{R}_+$, where $g: \mathbb{F} \times \mathcal{V} \rightarrow \mathbb{R}$, and $\mathbb{F}$ denotes the space of functions to which $\lambda$ belongs. The following assumption simplifies the dependence of the robot speed on the robot density.
\begin{asu} \label{asu_local_speed}
	Assume that the coarse-grained density of robots $\lambda$ varies slowly over the domain,
	\begin{equation}
	\left\lVert\frac{\partial \lambda}{\partial z} \right\rVert\ll 1, \forall z\in \mathcal{V}.
	\end{equation}
	Furthermore, let the speed of robot $i$ depend only on the densities in some neighborhood of the robot location $z_i$. Then, the speed of the robot can be assumed to depend only on the density at the robot's location, i.e., the function $g$ is simplified as
	\begin{equation}
	g(\lambda,z_i) \approx \lambda(z_i).
	\end{equation}
\end{asu}
Consequently, we denote the speed of robot $i$ as simply $v(\lambda(z_i))$. This assumption allows us to develop an analytical expression for the speed of a robot at a given location, as a function of the robot density at that location. Using the inter-robot collision model developed in our previous work~\cite{mayya2018localization}, the expected time between collisions experienced by robot~$i$ at its current location $z_i$ is given as,
\begin{equation} \label{eqn_tau}
\tau_{c}(\lambda(z_i))^{-1} = 4r\frac{4}{\pi}v_0\lambda(z_i),
\end{equation} 
where $r$ denotes the radius of each robot. The modified speed term $(4/\pi) v_0$ is equal to the mean relative speed between all the robots~\cite{mayya2018localization}. \par 
Furthermore, let $\tau_m$ denote the expected time spent by a robot in a collision with another robot before re-attaining its self-propelled speed $v_0$. This process occurs purely via forces acting on the robots as well as the rotational diffusion of each robot (as dictated by the rotational diffusion coefficient $D_r$ in \eqref{eqn_dyn}). For instance, the rotational diffusion might cause the robots to eventually move in different directions effectively resolving the collision. In Section \ref{sec:sim}, for a choice of diffusion parameters, we use a team of simulated brushbots to empirically compute the average speed and estimate $\tau_m$ as the parameter value which best fits the speed data. \par 
The robot is expected to travel at speed $v_0$ between collisions, and effectively remain stationary during the collision. Consequently, the average speed can be expressed as
\begin{equation} \label{eqn_vel_init}
v (\lambda(z_i)) = v_0 \left (1 - \frac{\tau_m}{\tau_c(\lambda(z_i)) + \tau_m}\right ).
\end{equation}
Under the simplifying assumption  that the time to resolve collisions is smaller than inter-collision time intervals, i.e., $ \tau_m \ll\tau_c$, (which is valid at intermediate densities), and substituting from \eqref{eqn_tau} the expression for $\tau_c$, \eqref{eqn_vel_init}, can be cast into the general form

\begin{equation} \label{eqn_vel}
v(\lambda(z_i)) = v_0\left (1 - \frac{\lambda(z_i)}{\lambda^*}\right ),
\end{equation}
where
\begin{equation}\label{eqn_lamb_star}
\lambda^* = \left (4r\frac{4}{\pi}v_0\tau_m\right )^{-1}
\end{equation}
represents the extrapolated density at which the speed becomes zero, and can be interpreted as the packing density of robots in the domain \cite{cates2015motility}. 
%\begin{equation} \label{eqn_vel}
%v(\lambda(z_i)) = v_0\Big (1 - \frac{\tau_m}{\tau_c(\lambda(z_i))}\Big)
%\end{equation}
Next, we introduce a simulation setup which verifies the linear dependency of velocity on local robot density (as predicted by \eqref{eqn_tau}, \eqref{eqn_vel} and \eqref{eqn_lamb_star}), and gives us numerical estimates for the collision resolution time~$\tau_m$. 
%As the robots move around the domain, the density of robots in the domain will vary. Let $\rho : D \rightarrow \mathbb{R}$ denote the density of robots as defined above. The robots interact with each other via excluded volume interactions alone. Thus, their speed varies based on the number of robots that they encounter. Let $v(\rho)$ denote the speed of a robot given the density in a region around it. \textcolor{red}{Specify that this is an approximation.}
\subsection{Simulation Setup} \label{sec:sim}
We validate the density-dependent speed model using a simulated team of brushbots operating in a closed and bounded rectangular domain $\mathcal{V}$. The simulation consists of $N$ disks of a given radius $r$ moving according to the dynamics described by \eqref{eqn_dyn}. During collisions, excluded volume constraints---a consequence of the fact that robots are not inter-penetrable---are imposed via a force acting on the robots, as is common in the physics literature, e.g., \cite{fily2014freezing}. In order to avoid the influence of boundary effects, periodic boundary conditions are applied, i.e., when robots exit from one side of the domain, they reappear on the other side. In Section \ref{sec:exp}, when performing experiments on the actual brushbots, we allow the robots to turn around when they reach the boundaries of the domain. \par
\begin{figure}
	\includegraphics[trim={3cm 0cm 2cm 1cm},clip,width=0.49\textwidth]{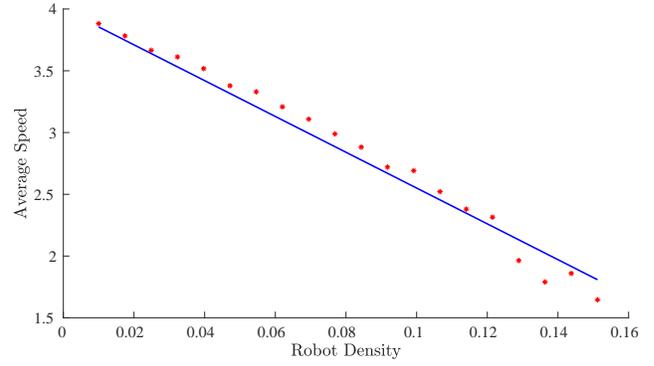}
	\caption{Validation of the relation between the average speed of robots and the swarm density using the simulation setup described in Section \ref{sec:sim}. Average speed measurements are denoted as red dots, while the best-fit line is denoted in blue. At low densities, the average speed equals the self-propelled speed of the robots (set at $v_0=4$ in this simulation). As the density increases, robots experience higher collision rates which slows them down. Best-fit collision resolution time $\tau_m = 0.177s$ (see \eqref{eqn_vel}). }
	\label{fig:vel}
\end{figure}

 
For varying robot densities, Fig. \ref{fig:vel} plots the speed of the robots averaged over all the robots. This simulation is performed for a uniform distribution of robots over the domain to prevent the averages from being affected by variable high and low density regions. A uniform distribution was ensured by selecting the activity parameter $\mathcal{A}$, defined in \eqref{eqn_activity}, well below the values which would lead to phase separation (see supplementary material for~\cite{stenhammar2013continuum}).  The average speed of the swarm at each point in time was computed by projecting the instantaneous velocity of each robot along its current orientation vector $\Theta_i = [\cos{\theta_i},\sin{\theta_i}]^T$,
\begin{equation}
\hat v(t) = \frac{1}{N}\sum_{i=1}^{N} \dot{z_i}(t)^T\Theta_i(t).
\end{equation}

As seen in Fig. \ref{fig:vel}, the average robot speed decreases linearly with density, as predicted by \eqref{eqn_tau} and \eqref{eqn_vel}. Deviations at high densities can be justified by the violation of the assumption that $\tau_m \ll \tau_c$. In the next section, we illustrate how the slowdown of robots due to collisions can actually lead to phase separated regions of high and low robot densities, as predicted by equilibrium thermodynamics.

\section{Motility-Induced Phase Separation} \label{sec:mips}
In the previous section, we analyzed the velocity profile of robots interacting purely via inter-robot collisions. In this section, we provide a brief summary of important results in the active matter literature which establish an equivalence between such a swarm of robots and a passive Brownian molecular system at equilibrium (see \cite{cates2015motility} for further details). This equivalence allows us to specify the system parameters under which the swarm can be expected to phase separate and form regions of unequal robot density. \par
\subsection{Theoretical Analysis} \label{sec:theo}
We begin the analysis by studying the probability distribution of a single particle whose speed, denoted as $v(\lambda(z))$, varies spatially over the domain according to the robot density. For a single robot obeying the Langevin dynamics in \eqref{eqn_dyn}, the probability density $\phi$ describing the location of a robot over the domain evolves according to the following equation, as given in \cite{cates2013active},
\begin{align}
&\dot \phi(z) = \nabla \cdot J, \\
&J = -D(\lambda(z))\nabla \phi(z) + V(\lambda(z))\phi(z), \forall z \in \mathcal{V},
\end{align}
where $\nabla \cdot$ denotes the divergence operator, $\nabla$ denotes the gradient, and $D$ is now the spatially varying diffusion coefficient of the robot, modified from \eqref{eq:eff_diff} as
\begin{equation}
D(\lambda(z)) = \frac{v^2(\lambda(z))}{2D_r} + D_t.
\end{equation}
The drift-velocity $V$ satisfies the relation
\begin{equation} \label{eqn_vd}
\frac{V(\lambda(z))}{D(\lambda(z))} = -\left (1+\frac{2D_tD_r}{v^2(\lambda(z))}\right )^{-1} \nabla \ln{v(\lambda(z))}.
\end{equation} 
From these equations, the coarse-grain density of robots interacting with each other has been shown~\cite{cates2015motility}
to evolve according to the following stochastic differential equation
\begin{equation} \label{eqn_den_de}
\dot \lambda(z) = -\nabla \cdot \Big(-D\nabla \lambda(z) + V\lambda(z) + \sqrt{2D\lambda(z)}\Lambda\Big),
\end{equation}
where $\Lambda$ represents white Gaussian noise interpreted in the It\^o sense \cite{cates2015motility}.
We suppress the explicit dependence of $D$ and $V$ on $\lambda(z)$ for readability. \par 
The steady state probability distribution of the density, $\mathcal{P}_\mathrm{eq}(\lambda)$, can be obtained by solving for the equilibrium solution of the corresponding Fokker-Planck equation, 
\begin{equation} \label{eqn_fp_eq}
\Bigg[V\lambda - D\nabla \lambda - D\lambda\Big(\nabla\frac{\partial}{\partial\lambda}\Big)\Bigg] \mathcal{P}_\mathrm{eq}(\lambda) = 0.
\end{equation}
The following theorem, initially presented in \cite{tailleur2008statistical} and summarized in \cite{cates2015motility}, illustrates how a swarm of robots interacting via collision interactions can be mapped to a system of passive Brownian particles at equilibrium.
\begin{thm}(\cite{cates2015motility}) \label{thm_ps}
	A team of robots operating in a domain $\mathcal{V}$, satisfying the dynamics in \eqref{eqn_dyn} and interacting via purely inter-robot collisions, can be expressed as a system of passive Brownian particles with an attractive potential if the following conditions are satisfied:
	\begin{enumerate}
		\item Assumption \ref{asu_local_speed} is valid.
		\item The coarse-graining of the microscopic density operator given by \eqref{eqn_mic_den} is valid and satisfies the dynamics given by \eqref{eqn_den_de}.
	\end{enumerate} 
	Under these conditions, the free energy density of the system can be expressed as,
	\begin{multline} \label{eqn_f_den}
	f(\lambda(z)) = \lambda(z)(\ln(\lambda(z))-1) ~+ \\
	\int_{0}^{\lambda(z)}\frac{1}{2}\ln{\Big (\frac{v^2(s)}{D_r}+ 2D_t\Big)}ds,~ \lambda(z) \leq \lambda^*.
	\end{multline}
	where $\lambda^*$ is given by \eqref{eqn_lamb_star}.
\end{thm} 
\begin{proofs}
	An equivalence between the self-propelled robot swarm and a passive Brownian particle system at equilibrium is made by expressing the steady-state solution of \eqref{eqn_fp_eq} as obeying the equilibrium Boltzmann distribution \cite{atkins2011physical} over the domain, given as
	\begin{equation} \label{eqn_boltz}
	\mathcal{P}_{eq} \propto exp(-\mathcal{F}),
	\end{equation}
	where $\mathcal{F} = \mathcal{F}_\mathrm{ent} + \mathcal{F}_\mathrm{vel}$ is the free energy of the system, expressed in terms of the free energy density functional $f$,
	\begin{equation}
	\mathcal{F} = \int_{z\in D} f(\lambda(z))\,dz.
	\end{equation}
	The free energy density, given by,
	\begin{equation}
	f(\lambda(z)) = \lambda(z)(\ln{\lambda(z)}-1) + f_\mathrm{vel}(\lambda(z)),
	\end{equation}
	comprises of two parts: an ideal entropy contribution (contributing to $\mathcal{F}_\mathrm{ent}$) and an excess energy density $f_\mathrm{vel}$ (contributing to $\mathcal{F}_\mathrm{vel}$). The latter, which would stem from the attractive potential in a passive Brownian system, in fact stems from the density dependent velocity profiles of the robots. To satisfy \eqref{eqn_boltz}, the following integrability condition results (discussed in \cite{tailleur2008statistical}),
	\begin{equation}
	\frac{V(\lambda(z))}{D(\lambda(z))}= -\nabla \frac{\partial \mathcal{F}_\mathrm{vel}}{\partial \lambda}.
	\end{equation}
	Substituting from \eqref{eqn_vd}, and observing that $\nabla \ln{v} = v^{-1}\nabla v$, this condition can be re-written as,
	\begin{equation} \label{eqn_int}
	\frac{v\nabla v\tau_r}{v^2\tau_r + 2D_t} = \frac{\partial \mathcal{F}_\mathrm{vel}}{\partial \lambda},
	\end{equation}
	where $\tau_r=D_r^{-1}$.
	Given Assumption \ref{asu_local_speed}, which implies that the speed of each robot only depends on the local density, \eqref{eqn_int} is satisfied by the following speed-derived free energy density
	\begin{equation}
	f_\mathrm{vel}(\lambda) = \int_{0}^{\lambda}\frac{1}{2}\ln{\Big (\frac{v^2(s)}{D_r}+ 2D_t\Big)}\, ds,
	\end{equation}
	under the constraint that the density $\lambda$ can never exceed the close packed value given by \eqref{eqn_lamb_star}. This leads to the desired result.
\end{proofs}
Given the equivalence established by Theorem \ref{thm_ps}, we can analyze the phase separation properties of the robot swarm by analyzing the free-energy density functional in \eqref{eqn_f_den}. The following observations summarize some theoretical characterizations of the conditions under which such a system can be expected to phase separate.
\begin{obs}
	In classical thermodynamics \cite{atkins2011physical}, spontaneous separation of a particle system into regions of high and low density occurs when concavities exist in the free energy density $f$ defined in \eqref{eqn_f_den}. More specifically, when the local density $\lambda(z)$ at a point $z\in \mathcal{V}$ is such that $f''(\lambda(z)) < 0$ where $'$ denotes a derivative with respect to the density $\lambda$, small fluctuations cause the system to phase separate into regions with densities which result in a reduction in free energy density. This process is called spinodal decomposition.
\end{obs}
Given \eqref{eqn_f_den}, it is a simple exercise to verify the conditions on the density-dependent velocity of the robots which favor occurrence of spinodal decomposition in the system \cite{cates2015motility},
 \begin{equation}
 f''(\lambda(z)) < 0 \iff \frac{v(\lambda(z))^2}{D_r}\left (1 + \lambda(z)\frac{v'(\lambda(z))}{v(\lambda(z))}\right ) < -2D_t.
\end{equation}
Substituting the expression of density dependent speed from \eqref{eqn_vel}, we can furthermore make the following observation regarding the \emph{conditions on robot density and parameters} under which the swarm can be expected to achieve non-uniform density distributions.

\begin{obs} \label{obs_ps}
	
	For a team of brushbots with velocity dependent speed given in \eqref{eqn_vel}, the spinodal densities (at which $f''(\lambda) = 0$), represented as $\lambda_s^{\pm}$, are given as
	\begin{equation}
	\lambda_s^{\pm} = \frac{\lambda^*}{4}\left ( 3 \pm \sqrt{1-16D_tD_r/v_0^2}\right ),
	\end{equation}
	where $	\lambda^*$ is specified in \eqref{eqn_lamb_star}. Consequently, these spinodal points only exist when the following condition is satisfied by the robot parameters: 
	\begin{equation} \label{eqn_vel_con}
	v_0 > \sqrt{16D_tD_r}.
	\end{equation}
	Since robot densities cannot exceed the packing density $\lambda^*$, the following ranges of densities, 
	\begin{equation} \label{eqn_den_range}
	\lambda_s^- \leq \lambda(z) \leq \min(\lambda_s^+,\lambda^*),\quad z \in \mathcal{V}
	\end{equation}
	favor spontaneous phase separation. That is, when the local density in a region of the domain lies between these values, the swarm can be expected to spontaneously phase separate into regions with low and high robot densities. 
\end{obs}
\subsection{Simulations} \label{sec:sim_mips}
\begin{figure}
	\subfloat[$t = 0s$]{
		\includegraphics[trim={3cm 1cm 2.cm 0.8cm},clip,width=0.49\columnwidth]{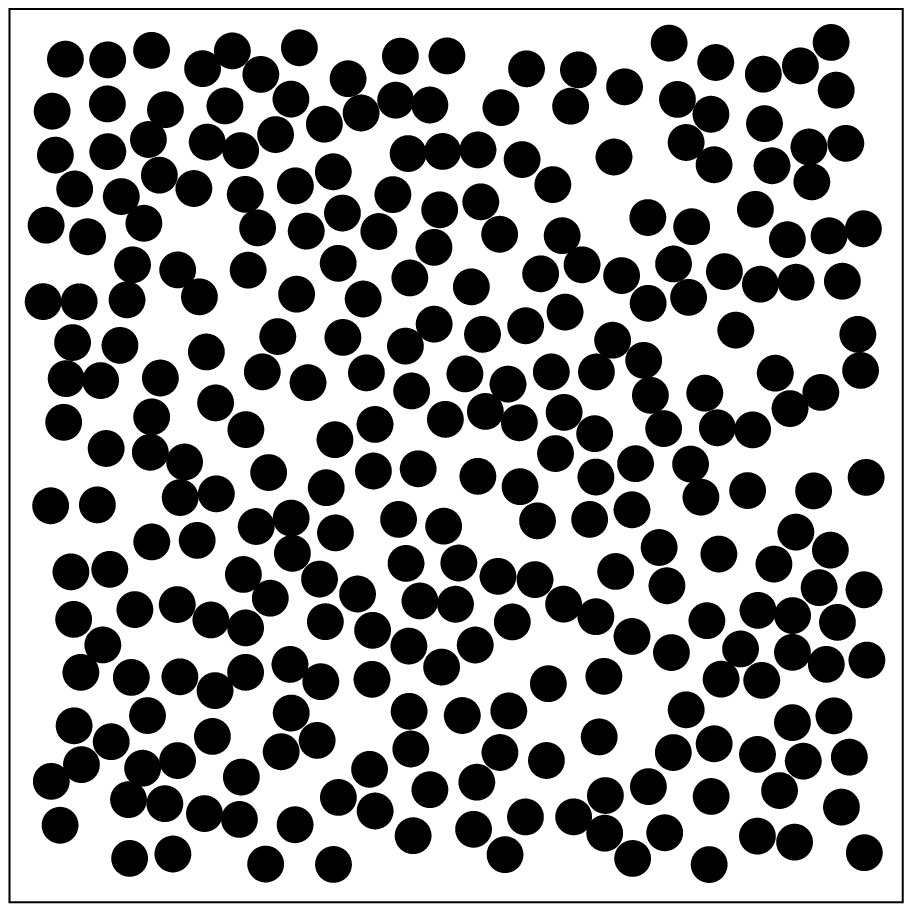}
		\label{mips_a}
		%\caption{t}
	}
	\subfloat[$t = 15s$]{
		\includegraphics[trim={3cm 1cm 2.cm 0.8cm},clip,width=0.49\columnwidth]{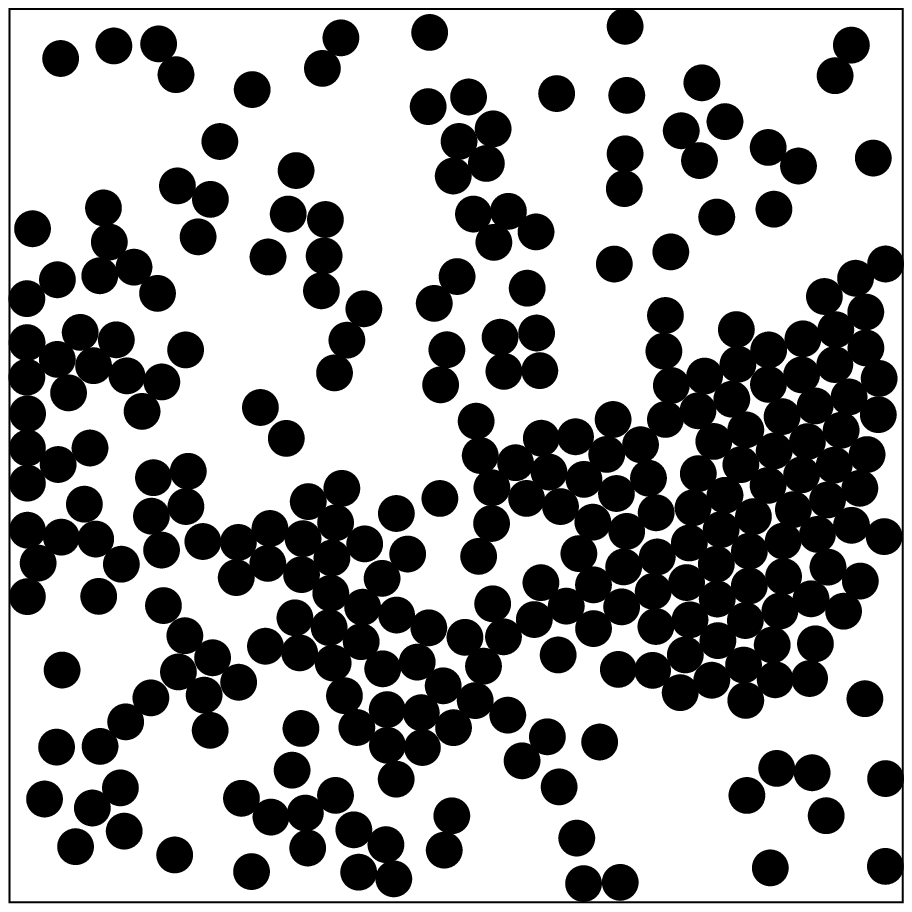} 
		\label{mips_b}	%t=15	
	}
	\caption{Snapshots illustrating the formation of high as well as low density regions concurrently in a team of simulated robots operating in a domain with periodic boundary conditions. For a set of simulation parameters ($N = 292,r=1,v_0 = 4,D_r=1e^{-4},D_t=1e^{-5}$), each snapshot represents the configuration of the system at different times. Inter-robot collisions slow down the robots, which cause additional robots to join the clusters. This leads to the formation of dynamic high-density clusters along with the existence of lower density regions in the domain.}
	\label{fig:snaps}
\end{figure}

We observe the formation of dynamic robot clusters in a rectangular domain with periodic boundary conditions using the simulation setup described in Section \ref{sec:sim}. Fig.~\ref{fig:snaps} illustrates the intermediate-time snapshots of the phase separating swarm robotic system. The robots are placed according to a uniform distribution in the domain at time $t=0$ (see Fig. \ref{mips_a}) and simply move according to the Langevin dynamics described in Section \ref{sec:mm} while experiencing collisions among each other. Figure~\ref{mips_b} depicts how a uniformly distributed swarm of robots can form regions of high and low robot density, primarily caused by the density-dependent slow down of robots which precipitates the formation of dynamic high-density robot clusters. \par 
\begin{figure}[t]
	\includegraphics[trim={1cm 0.1cm 1cm 0.8cm},clip,width=\columnwidth]{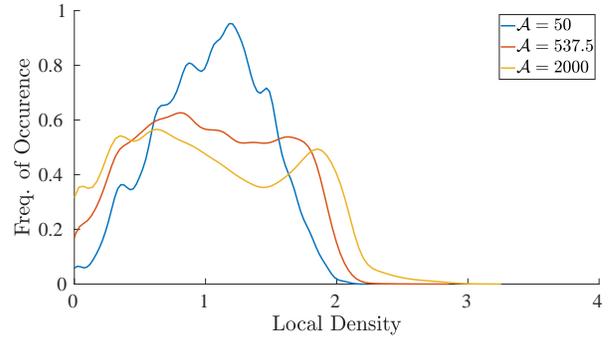}
	\caption{Empirically obtained distribution of robot densities evaluated over a grid of size $l^2$ with $l=10$. A Gaussian kernel smoothing was applied on the histogram of robot densities collected over multiple simulations at a constant robot density (simulation parameters: $r=1,D_t=1e^{-5},N=382$). As seen, for high activity parameters, the distribution is distinctly bimodal, due to the formation of high and low robot density regions as predicted in Section \ref{sec:mips}. For lower activity parameters, the swarm does not phase separate and the density distribution over the grid remains unimodal. This presents a mechanism to characterize the simultaneous existence of higher and lower robot densities in the domain. The displayed data was collected over multiple simulation runs.}
	\label{fig:den_dist}
\end{figure}

We characterize the simultaneous existence of regions with higher and lower robot densities by computing the coarse-grained density $\lambda$ over the domain,
\begin{equation}
\lambda(z) = \sum_{i=1}^{N} w(\|z - z_i\|), \forall z \in \mathcal{V},
\end{equation}
where the weighting function $w$ is given as,
\begin{equation}
w(d) = \text{exp}\big(-d_c^2/(d_c^2-d^2)\big).
\end{equation}
Here, $d_c$ is the cutoff distance at which $w(r) \rightarrow 0$. In simulation, the coarse-grained density was evaluated on a grid of $l^2$ lattice points, with cutoff distance $d_c = 0.8l$. For a grid size $l = 10$, Fig. \ref{fig:den_dist} plots the empirically obtained distribution of the coarse-grained densities evaluated at the lattice points. These values were obtained by applying a Gaussian kernel-smoothing on the histogram of coarse-grained densities over the grid. For each set of activity parameters, data points were collected from multiple simulations of the robot swarm, to average out effects due to initial conditions. As seen, the distribution of robot densities is unimodal at low activity levels and becomes distinctly bimodal at higher activities, indicating the formation of low and high robot densities in the domain. \par
\begin{figure}[h]
	\includegraphics[trim={1cm 0.1cm 1cm 0.8cm},clip,width=\columnwidth]{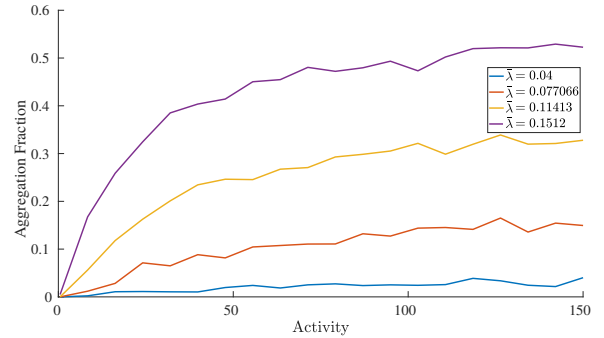}
	\caption{Dependence of the average aggregation fraction on the activity parameter of the robots (simulation constants:~$r = 1, D_t = 1e^{-5}$). The aggregation fraction measures the fraction of robots which belong to high-density clusters (chosen with a cut-off size $N_c=4$) averaged over time. As seen, at low activity levels, the extent of aggregation remains low regardless of the density of robots in the domain. As the activity is increased, the fraction of robots in high density clusters increases. $\bar \lambda$ denotes the mean density of robots in the domain.}
	\label{fig:activity_ps}
\end{figure}

\begin{figure}[h]
	\includegraphics[trim={1cm 0.1cm 1cm 0.8cm},clip,width=\columnwidth]{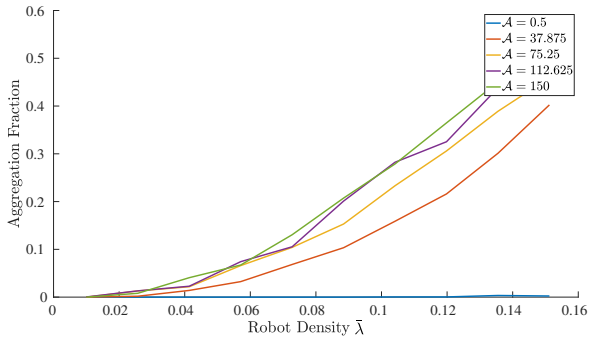}
	\caption{Dependence of the aggregation behaviors on the mean density of robots in the domain, as predicted by \eqref{eqn_den_range}. Below a certain density, no significant aggregation is seen in the robots regardless of the activity parameter of the robots (simulation constants:~$r = 1, D_t = 1e^{-5}$). This is primarily because robots are able to resolve collisions (therefore dissolving clusters) before other robots can join the aggregation.}
	\label{fig:density_ps}
\end{figure}

To quantify the formation of clusters in the swarm, we measure the time-averaged fraction of robots that belong to high-density aggregations in the swarm. An aggregation is defined as robots in physical contact with each other, whose total size exceeds a minimum cut-off value $N_c$. Figure \ref{fig:activity_ps} illustrates the impact of the activity parameter $\mathcal{A}$ of the robots, by plotting the average aggregation fraction for increasing activity values and for different mean robot densities in the domain (denoted as $\bar \lambda = N/|\mathcal{V}|$ where $|\mathcal{V}|$ denotes the area of domain $\mathcal{V}$). As seen, no significant robot aggregation occurs below a certain activity threshold of the robots, regardless of the density. This observation is in agreement with the lower activity thresholds for observed phase separation in major studies, e.g., \cite{cates2015motility}. Figure \ref{fig:density_ps} illustrates the need for a sufficient robot density to see phase separation behaviors, as predicted by \eqref{eqn_den_range}. \par

\section{Deployment and Experiments} \label{sec:exp}
In this section, we first identify the noise parameters of a single brushbot and discuss the implications of these parameters on the phase separation properties of the swarm. Following this, we illustrate the emergence of dynamic regions of low and high robot density using a team of 26 brushbots operating in a confined rectangular environment. \par 
In order to identify the translational and rotational diffusion coefficient of the brushbots, we excited one brushbot with a constant self-propelled linear speed and zero angular velocity in a confined rectangular space. Position and orientation data was collected using an overhead tracking system. When a robot encountered the walls of the domain, the data collection was terminated until the robot turned around to traverse the environment again. Data was collected for a total time of $300$s at a self-propelled speed $v_0 = 6cm/sec$. \par 
The rotational diffusion coefficient can be computed by measuring the mean-square variation in the orientation of a robot \cite{marino2016dynamics},  
\begin{equation}
D_r = \frac{1}{2}\lim_{t\to \infty} \frac{d}{dt}\big\langle\theta(t)-\theta(0)\big\rangle^2,
\end{equation}
where $\big\langle\theta(t)-\theta(0)\big\rangle^2$ represents the mean-square orientation change in a time duration $t$, averaged over multiple experimental runs. We numerically estimate $D_r$ by performing a linear regression on the mean-square variations in the orientation of the robot for varying time-intervals (estimated value $D_r = 0.0041$). Noticeably, the value for the translational diffusion coefficient $D_t$ as computed from these experiments was negligible within numerical precision (computed using the Green-Kubo method \cite{marino2016dynamics}). \par 
These estimates of the noise characteristics of a brushbot, allows us to make the following observation regarding the ability of a swarm of brushbots to spontaneously phase separate
\begin{obs}
Based on the robot noise parameters identified ($D_r = 0.014, D_t = 0$), \eqref{eqn_vel_con} is satisfied always. Consequently, a swarm of brushbots can be expected to spontaneously form regions of low and high robot density as long as the number of robots are high enough to achieve the local densities described in \eqref{eqn_den_range}. 
\end{obs}
It should be noted that the identification of diffusion parameters for the brushbots performed above is not meant to serve as a quantitative analysis of the noise properties of the brushbots, but to understand the qualitative effects of these values on the formation of lower and higher density regions in the swarm. Indeed, since the effects from translational noise are minimal, the exact values of the noise parameters do not play a role in the prediction of motility-induced phase separation. \par
\begin{figure*}[t]
	\includegraphics[trim={0cm 0.0cm 0cm 0.0cm},clip,width=0.5\columnwidth]{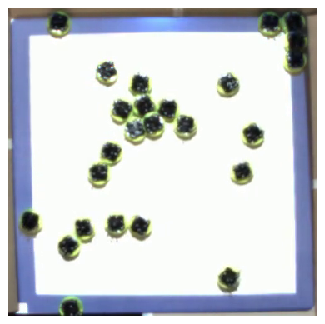}~%
	\includegraphics[trim={0cm 0.0cm 0cm 0.0cm},clip,width=0.5\columnwidth]{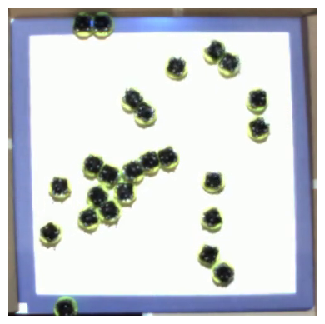}~%
	\includegraphics[trim={0cm 0.0cm 0cm 0.0cm},clip,width=0.5\columnwidth]{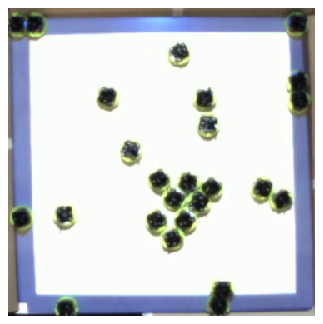}~%
	\includegraphics[trim={0cm 0.0cm 0cm 0.0cm},clip,width=0.5\columnwidth]{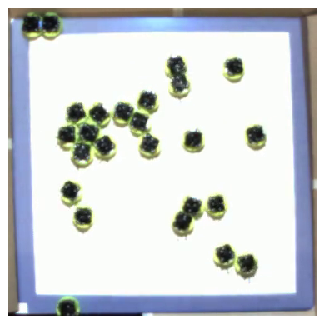}
	\caption{Snapshots for a team of 26 brushbots propelled at a constant speed in a domain. Reflective boundary conditions were applied by injecting an angular velocity to the robots when they hit the boundaries. As predicted by phase separation theory, collisions cause the robots to slow down which leads to the formation of high density robot clusters which co-exist along with regions of lower robot density. These clusters form and dissolve over time in different regions of the domain.}
	\label{fig:exp_data}
\end{figure*}

 
Figure \ref{fig:exp_data} illustrates intermediate-time snapshots of 26 brushbots achieving regions of varying robot density in an enclosed square domain. A constant speed and zero angular velocity input was supplied to the robots. Specular reflection boundary conditions were imposed by injecting the robots with an angular velocity when they hit the boundary of the domain, reorienting them towards the interior of the domain. \par
\begin{figure}[h]
	\includegraphics[trim={1cm 0.1cm 1cm 0.8cm},clip,width=\columnwidth]{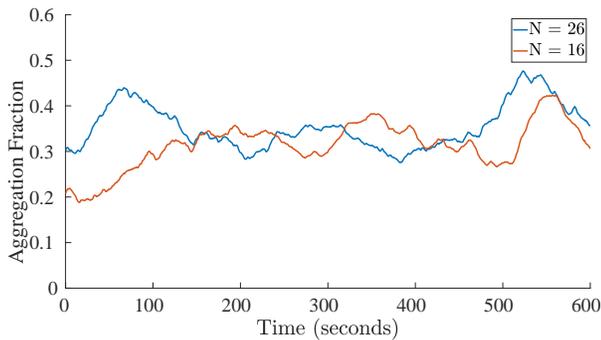}
	\caption{Fraction of robots belonging to high density robot aggregations for a team of real brushbots. An aggregation is defined as a collection of robots beyond a cut-off size in physical contact with each other. The robots travel randomly in the domain while colliding with each other. The reduction in speed caused due to collisions precipitates the formation of simultaneous regions of low and high robot density as predicted by the phase separation theory in Section \ref{sec:mips}. As seen, the fraction of robots in aggregations remains fairly significant throughout the experiments.}
	\label{fig:exp_agg}
\end{figure}

As expected, collisions caused the robots to slow down, which precipitated the formation of high density regions.
Robot clusters were identified based on physical contact among the robots, beyond a minimum threshold size (chosen as 4). We quantitatively analyzed the formation of these clusters by plotting the aggregation fraction in the swarm over time. This represents the fraction of robots which belong to a cluster at any given point in time. As seen in Fig. \ref{fig:exp_agg}, at least $20\%$ of the robots were clustered in aggregations at any point of time, while the rest of the robots moved freely through the domain.

\section{Conclusions}\label{sec:conc}
This paper demonstrates a mechanism to achieve regions of non-uniform density on a team of brushbots which possess no sensors to detect other robots but simply traverse the environment while colliding with each other. Analysis of similar self-propelled particle systems in the physics literature provide a theoretical basis for the the formation of higher and lower robot density regions. We characterize the average speed a robot as a function of the swarm density, and illustrate that certain conditions on this speed profile can lead to distinct regions of robot density. This emergent behavior was demonstrated on a team of 26 brushbots operating in a confined space while colliding with each other. 
%\clearpage

%%%%%%%%%%%%%%%%%%%%%%%%%%%%%%%%%%%%%%%%%%%%%%%%%%%%%%%%%%%%%%%%%%%%%%%%%%%%%%%%

\bibliographystyle{IEEEtran}

\bibliography{references}

\end{document}